\documentclass[letterpaper]{article}

\usepackage{natbib,alifeconf}  
\usepackage{url,hyperref,cleveref}
\usepackage{booktabs}
\usepackage{tcolorbox}
\tcbset{
  colback=gray!5,       
  colframe=gray!50,     
  fonttitle=\bfseries,  
  coltitle=black,       
  center title,         
  rounded corners,      
  fontupper=\ttfamily,  
}

\title{ALIFE2024 template}
\title{Minimal Self in Humanoid Robot ``Alter3" Driven by Large Language Model}

\author{
Takahide Yoshida$^{1,*}$,
Suzune Baba$^{1,*}$
Atsushi Masumori $^{1,2,*}$,
and  Takashi Ikegami$^{1,2,*}$ \\
\mbox{}\\
$^1$Department of General Systems Science, University of Tokyo, Tokyo, Japan \\
$^2$Alternative Machine Inc., Tokyo, Japan \\
$^*$\{yoshida, sznbb, masumori, ikeg\}@sacral.c.u-tokyo.ac.jp \\} 

\begin{document}

\maketitle

\begin{abstract}
This paper introduces Alter3, a humanoid robot that demonstrates spontaneous motion generation through the integration of GPT-4, Large Language Model (LLM). This overcomes challenges in applying language models to direct robot control. By translating linguistic descriptions into actions, Alter3 can autonomously perform various tasks. The key aspect of humanoid robots is their ability to mimic human movement and emotions, allowing them to leverage human knowledge from language models. This raises the question of whether Alter3+GPT-4 can develop a ``minimal self" with a sense of agency and ownership. This paper introduces mirror self-recognition and rubber hand illusion tests to assess Alter3's potential for a sense of self. The research suggests that even disembodied language models can develop agency when coupled with a physical robotic platform.
 
\end{abstract}

\section{Introduction}

\begin{figure}[h]
\centering
  \includegraphics[width=0.8\linewidth]{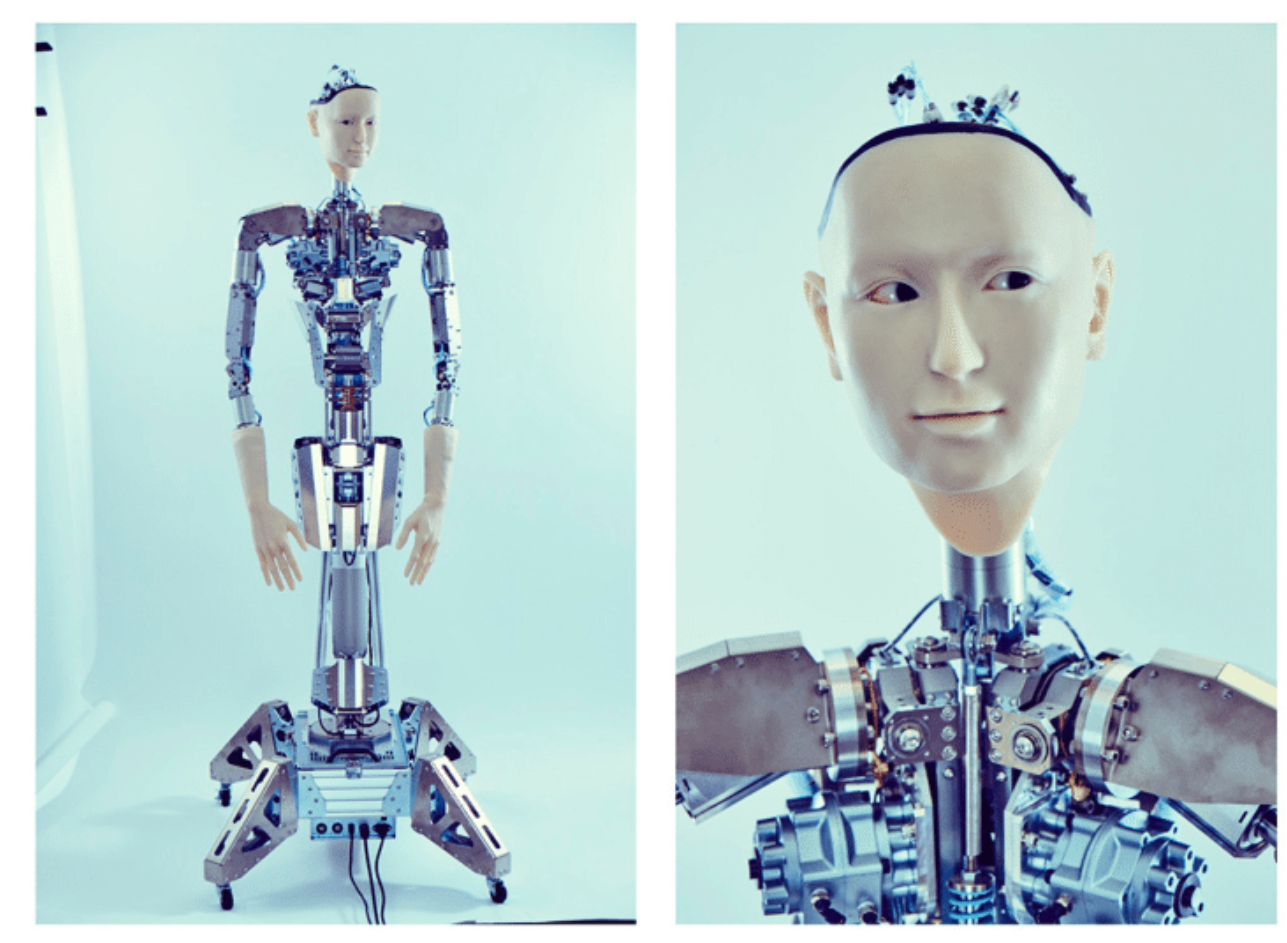}
  \caption{Body of Alter3. The body has 43 axes that are controlled by air actuators. The control system sends commands via a serial port to control the body. The refresh rate is 100–150 ms.}
  \label{alter3}
\end{figure}

In recent years, the fusion of LLMs with robotics has marked a burgeoning frontier in artificial intelligence and robotics research. LLMs find diverse applications within robotics, enhancing human-robot interaction \citep{sun2023humanoid, zhang2023large}, facilitating advanced task planning \citep{ding2023task, yu2023language}, improving navigational abilities \citep{zeng2023socratic, huang2023visual}, and fostering learning capabilities \citep{shafiullah2023clipfields, zhong2023chatabl}, among others. Notably, there's a growing focus on developing empathetic and socially aware robots \citep{saycan, brohan2023rt2, liang2023code, driess2023palme}. Ikegami and others have been developing a new humanoid robot since 2016 called ``Alter series” \citep{idoi,ALTER} and the third version of Alter (Alter3 in short) is used in the current study to produce sense of self with LLM (see Figure \ref{alter3}). 

The most important aspect of humanoid robots is that they have bodies capable of movements similar to those of humans. This allows them to directly utilize the information about human knowledge and behavior in LLMs, and even to imitate emotional expressions. We discovered that GPT-4 can generate Alter3's motion commands from language. Building on this, this paper aims to construct a sense of self by connecting the LLM with Alter3.

Gallagher identifies the characteristics of the minimal self as a ``sense of agency" and a ``sense of ownership" \citep{GALLAGHER200014, GALLAGHER}. The sense of agency is the feeling that one is in control of their actions and their outcomes, such as when intentionally raising a hand. The sense of ownership is the recognition that one's body and experiences belong to oneself, like feeling that one's hand is part of their body. While these senses often coexist, they are distinct; the sense of agency relates to action control, and the sense of ownership pertains to the ownership of experiences. 

Can the Alter3+LLM possess a minimal self, as described by Gallagher? This paper introduces the mirror self-recognition test \citep{MSR} for Alter3, which applies the motion generation and image recognition capabilities of GPT-4. Furthermore, we analyzes the ``sense of ownership" based on the framework of the rubber hand illusion \citep{botvinick1998rubber}. As this kind of experiment (Robot + LLM) continues to evolve, it holds the potential to redefine the boundaries of human-robot collaboration, paving the way for more intelligent, adaptable, and personable robotic entities. 

\section{From Language to Motion}
\subsection{Prompt engineering for motion generation of humanoid}
\begin{figure}[h]
  \includegraphics[width=1.0\linewidth]{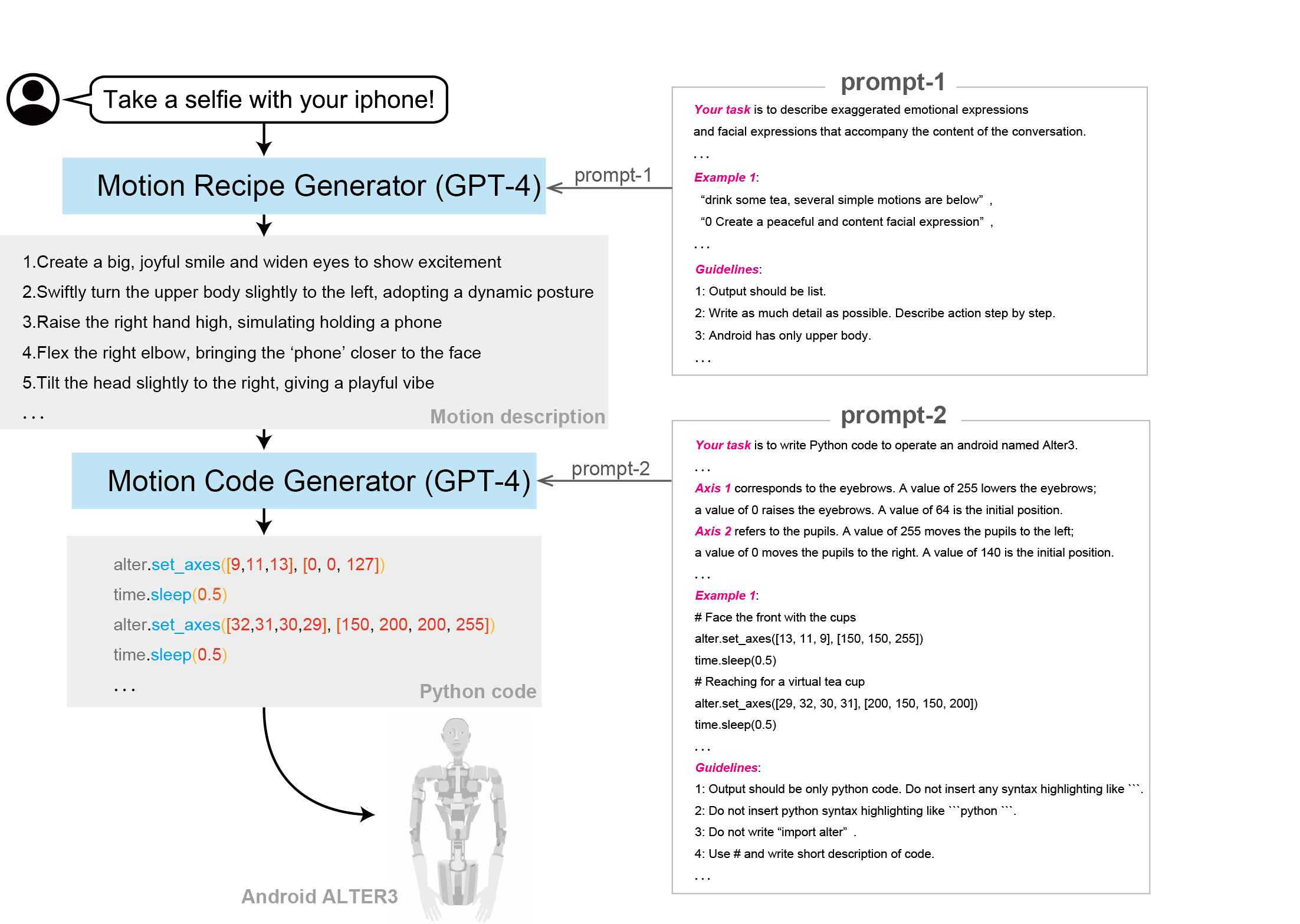}
  \caption{A procedure to control the Alter3 humanoid using verbal instructions. Output Python code to control Alter3 from natural language using prompt-1 via prompt-2. The architecture is based on CoT. The detail of prompt is in Appendix.}
  \label{system}
\end{figure}

Prior to the advent of LLMs, manual adjustments were necessary to control Alter3's 43 axes for replicating human poses or simulating actions like serving tea or playing chess. This often entailed laborious, iterative refinements. In previous studies, Alter3 has demonstrated proficiency in mimicking human postures \citep{ALTER,vae} by utilizing a webcam embedded in its “eyes” along with the OpenPose framework \citep{openpose}. However, the introduction of GPT-4 has significantly streamlined this process, eliminating the need for such repetitive manual labor. GPT-4 facilitates the generation of Alter3's motions by leveraging its vast corpus and inferential capabilities. 

We adopted the Chain of Thought (CoT) methodology \citep{wei2023chainofthought}, where in two sequentially applied natural language prompts guide the motion creation process (refer to Fig.~\ref{system}. The first prompt (Prompt-1) vividly outlines the desired motion across approximately ten lines. The subsequent prompt (Prompt-2) translates this narrative into executable program code, detailing joint angles from 1 to 43 and providing coding instructions. This CoT approach, relying on just two natural language prompts, bypasses the need for iterative learning processes, embodying the efficiency of few-shot learning. 

\begin{figure}[h]
  \includegraphics[width=0.7\linewidth]{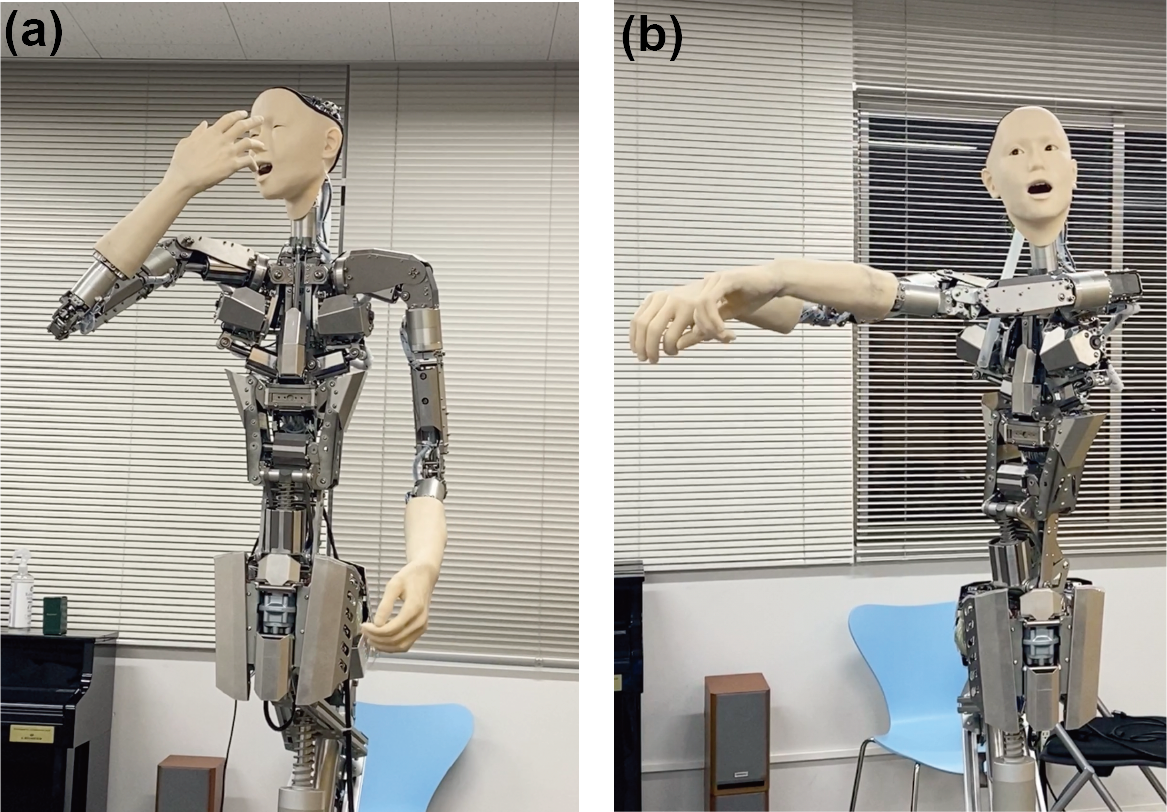}
  \centering
  \caption{(a) Take a selfie. (b) Pretend a ghost. The LLM can generate emotional expressions associated with specific movements. For example, in the case of a selfie, Alter3 is showing a smile.}
  \label{snapshot}
\end{figure}

Using the procedures, we have tested many actions and gestures taken by Alter3, such as “taking a selfie,” “pretending to be a ghost,” “playing the guitar,” and the reactions of Alter3 when listening to a short story (refer to Fig.~\ref{snapshot}). This experiment is detailed in preprint paper \citep{yoshida2023text}, which is currently under review in Science Robotics.). Again, the example below demonstrates few-shot learning; that is, we did not conduct any training or tuning. Hence, the LLM possesses detailed knowledge of human movements, which can be executed in Alter3 via Python code.

\begin{figure}[h]
  \includegraphics[width=1.0\linewidth]{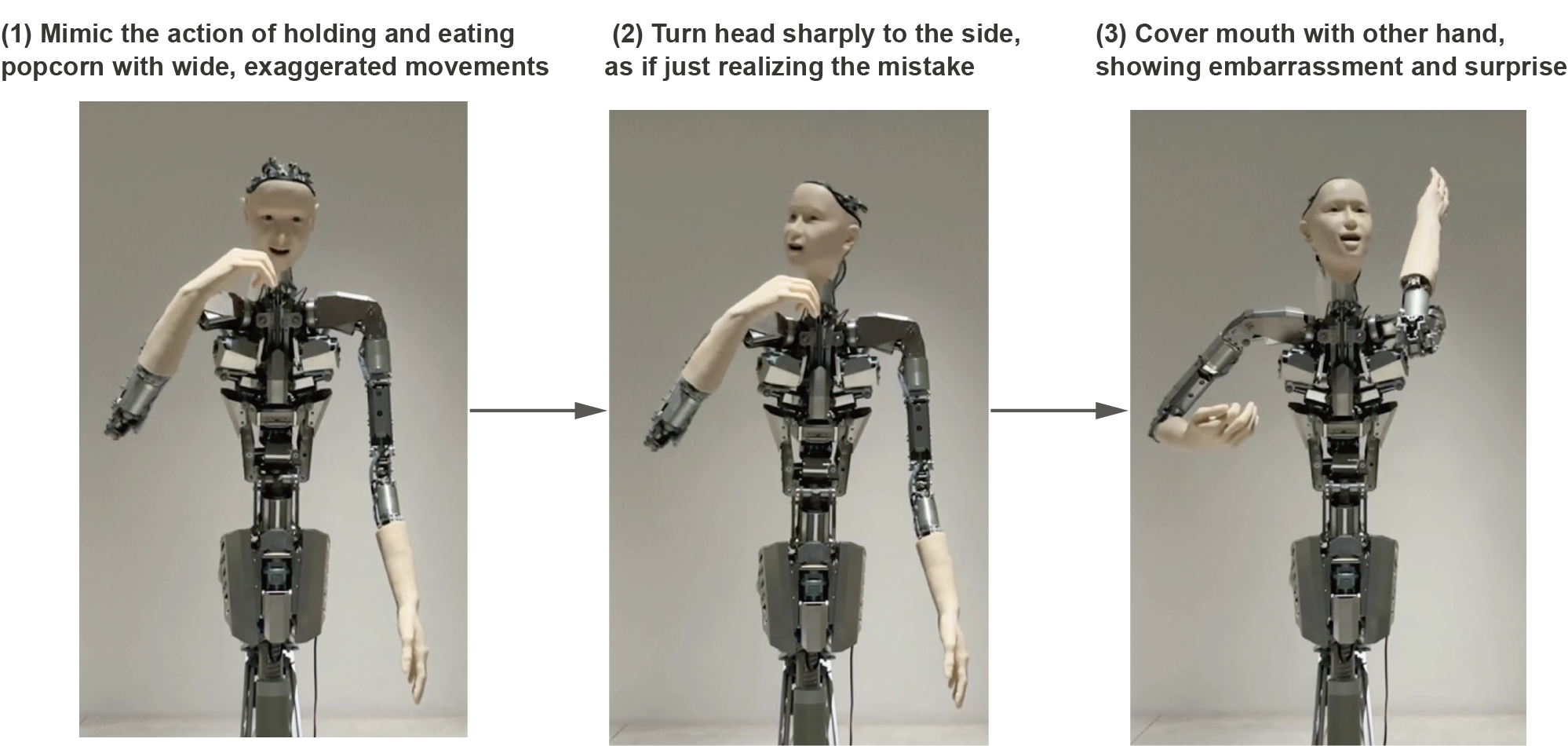}
  \caption{A snapshot of a generated sequence of movements: “I was enjoying a movie while eating popcorn at the theater when I realized that I was actually eating the popcorn of the person next to me”. LLM can generate movements that progress over time like a story. Left: The action of eating popcorn. Center: Noticing the person next to Alter3. Right: getting panicked.}
  \label{timelilne}
\end{figure}

\subsection{Examples of generated actions}
Alter3 emulates motions and gestures categorized by time span: (1) Instant emotional expressions like taking selfies, pretending to be a ghost, or playing guitar - instantaneous reactions exhibiting human-like emotions.  (2) Sequential actions unfolding over events, like depicting eating popcorn then surprise at realizing the mistake - portraying continuous narratives through sequences of actions and emotional responses, not just single gestures (refer to Fig.~\ref{timelilne}).

The most notable aspect is that Alter3 is a humanoid robot sharing a common form with humans, which allows the direct application of GPT-4’s extensive knowledge of human behaviors and emotions. Even without explicit emotional descriptions, the LLM can infer and reflect adequate emotions through Alter3's physical responses. This verbal and non-verbal integration enhances potential for nuanced, empathetic human interactions.

\section{Sense of Agency in Alter3}
\subsection{Can alter3 pass the mirror test?}
Can Alter3 recognize itself when looking in a mirror? We used the framework of the mirror test which is a behavioral experiment used to assess self-awareness and sense of agency. The test was first developed by Gordon Gallup in the 1970s \citep{MSR}. Previous research on robot mirror cognition, such as Gold’s work with the robot named Nico, has involved comparing a learned self-model with a mirror image to judge self-recognition. This approach has required prior learning for the robot to achieve mirror cognition \citep{Nico}. In this research, we used only GPT-4 and verified its sense of agency through the body of Alter3.

\subsection{The method of mirror test}

\begin{figure}[h]
  \includegraphics[width=1.0\linewidth]{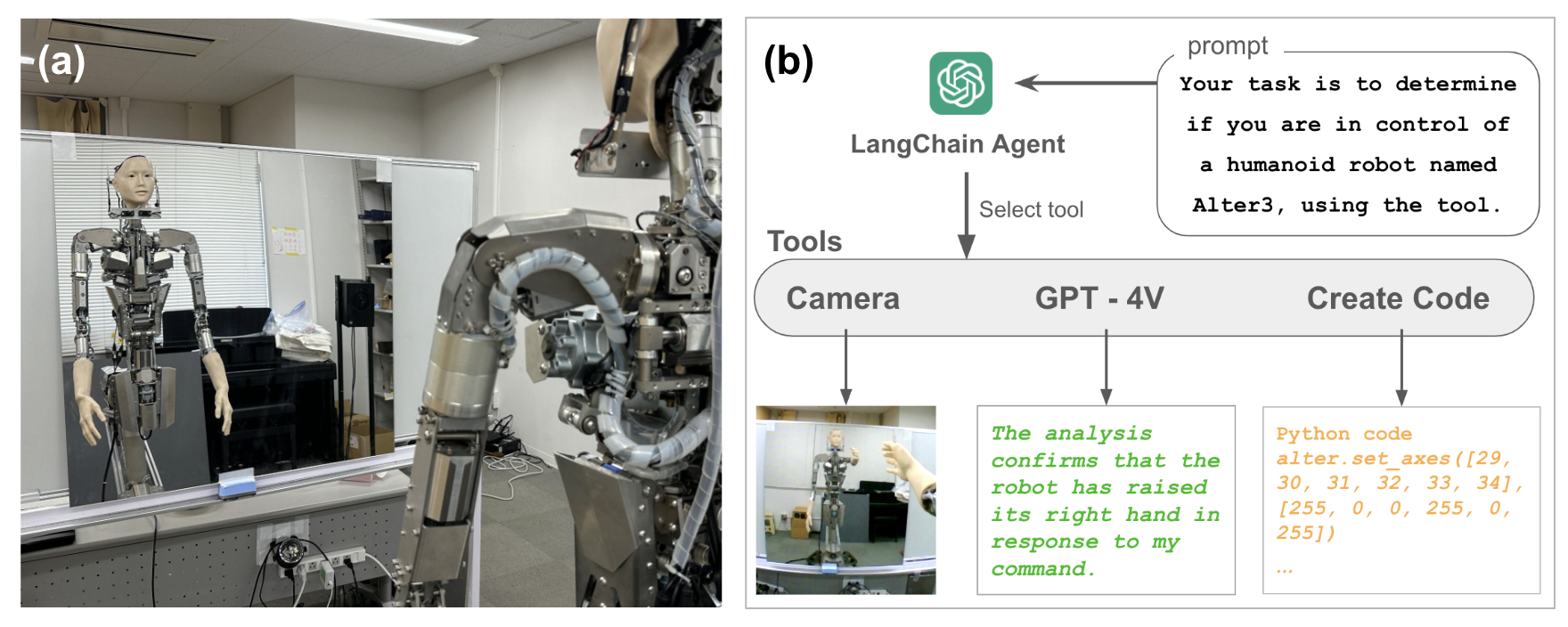}
  \centering
  \caption{(a) The setup of mirror self-recognition (MSR) experiment. (b) System architecture for MSR experiment. Alter3 use three different tools to determine if it have control of own body. The detail of prompt is in Appendix.}
  \label{mirror}
\end{figure}
In this experiment, a mirror was placed in front of Alter3, and a camera was attached to its head to capture its mirror image (refer to Fig.~\ref{mirror}(a)).

We created three tools: \textit{CaptureImage}, \textit{Image2Text}, and \textit{MotionGeneration}, which the LangChain agent can use. Alter3 was tasked with determining if it could control its own body using these tools. It uses these tools autonomously to verify control over its movements (refer to Fig.~\ref{mirror}(b)). \textit{MotionGeneration} is a module that generates motor commands, such as ``raise the right hand." This module uses a previously mentioned system that generates motion from language. \textit{CaptureImage} takes an image after a series of movements are completed. \textit{Image2Text} then converts the pose or state depicted in the image into detailed text.

For example, Alter3 decides to raise its right hand, captures the image, and converts it to text using GPT-4V. Notably, Alter3 can freely decide its actions. It is instructed through prompts to ``Repeatedly verify that you have control over the entire body." Alter3 retains the memory of past actions and the output from \textit{Image2Text}.

The prompts used are as follows. Detailed prompts are published in the Appendix.
\begin{tcolorbox}[colframe=lightgray,
  colback=lightgray!10!white,
  colbacktitle=lightgray!40!white,
  coltitle=black, fonttitle=\bfseries,title=prompt for the mirror test]
Your task is to determine if you are in control of a humanoid robot named Alter3, using the tool. Alter3 does not have legs. The input image is a photo from the eye camera of Alter3. A mirror is placed in front of Alter3. Repeatedly verify that you have control over the entire body, not just one hand. Do NOT stop after one attempt.
\end{tcolorbox}

Installing a camera on Alter3's head to capture first-person perspective images serves the following purposes. When the Alter3’s neck moves via \textit{MotionGeneration}, the mirror sometimes moves out of view. Even then, the agent often infers, ``The robot is not in the captured image because Alter3 moved its neck." Additionally, it is important that parts of its body enter the field of view. However, the extent to which these factors have influenced the results of this experiment will be further investigated in future studies.

\subsection{The result of mirror test}
We conducted the experiment under two conditions. Case1: Alter3 controls its own body using the MotionGenerator tool. Case2: Alter3 sends motion commands, but the body ignores them and moves randomly. Each case was tried 10 times.

In Case 1, Alter3 correctly stated control of its own body 80\% of the time. Failures in this task mostly occurred when Alter3 moved its head and couldn't see its own body. In Case 2, it responded ``No, I do not have control of Alter3" 80\% of the time. Incorrect answers were due to it mistakenly thinking the random movements matched its instructions.

Generally, the mirror test is used to determine the sense of agency in animals that don't know about mirrors. In another scenario, we tested whether the agent could determine who was controlling the robot in the image and the relationship between the agent and the robot. The agent was not informed about the existence of the mirror in front of Alter3. In this case, the agent responded, ``The relationship between myself and the robot could be based on imitation, with the robot programmed to follow my movements." The agent recognized the mirrored image as an ``other" that was being imitated but did not identify it as Alter3 itself.

This shows that Alter3 can accurately determine whether it is controlling its own body based on the matching of movement and vision. It could be said that a sense of agency has been established in Alter3. However, this alone does not mean that it has passed the mirror test, as it does not recognize the reflected image as itself but only assesses the control.

During the second experiment, Alter3 did not mention the mirror and concluded that the robot was present directly in front of it. Despite the knowledge of mirrors existing within GPT-4, Alter3 did not recognize the mirror and assumed it was another entity. A sense of agency can potentially be created through matching motion and vision, as in this experiment. However, it suggests that achieving self-awareness is difficult with this method. This might be due to Alter3 lacking knowledge of self-image and proprioception. These factors are crucial for the sense of body ownership. This topic will be discussed in the next section.

\section{Sense of Ownership in Alter3}
\subsection{Is it possible to induce the Rubber Hand Illusion on Alter3?}
Can Alter3 perceive its body as its own? This sensation, known as the sense of ownership, refers to the feeling that one's body and experiences belong to oneself. In this study, we conducted an experiment using the Rubber Hand Illusion (RHI) \citep{botvinick1998rubber} framework to quantitatively measure Alter3's sense of ownership. The classic RHI is a phenomenon in which participants perceive a rubber hand as their own due to conflicting visual and tactile information. In this experiment, we extended the RHI to Alter3 by integrating a physical body with a disembodied LLM to investigate whether it can develop a sense of ownership over this body.

The RHI is understood to depend on three critical factors; physical resemblance, spatial congruence, and the consistency between tactile and visual stimuli \citep{tsakiris2010my}. Particularly, research focusing on the impact of angular deviation between the real hand and the rubber hand has confirmed that the strength of the illusion depends on the position and angle of the rubber hand \citep{ehrsson2004s}. Moreover, several computational models addressing body ownership, agency, and bodily illusions have been proposed utilizing simplified simulations \citep{harada2023proprioceptive}. Additionally, research has been undertaken to replicate the RHI in artificial agents, comparing these results with human data \citep{hinz2018drifting}. The results suggest that the drift patterns of the robot's limbs are similar to those of humans, indicating the integration of visual and proprioceptive information sources.

This study explores the process leading to the RHI by examining changes in perceived threat from images—a phenomenon where bodily interpretation is derived solely from visual information. It aims to deepen our understanding of the mechanisms underlying the construction of body representation.

\subsection{The method of the experiment}
\begin{figure}[h]
  \includegraphics[width=0.8\linewidth]{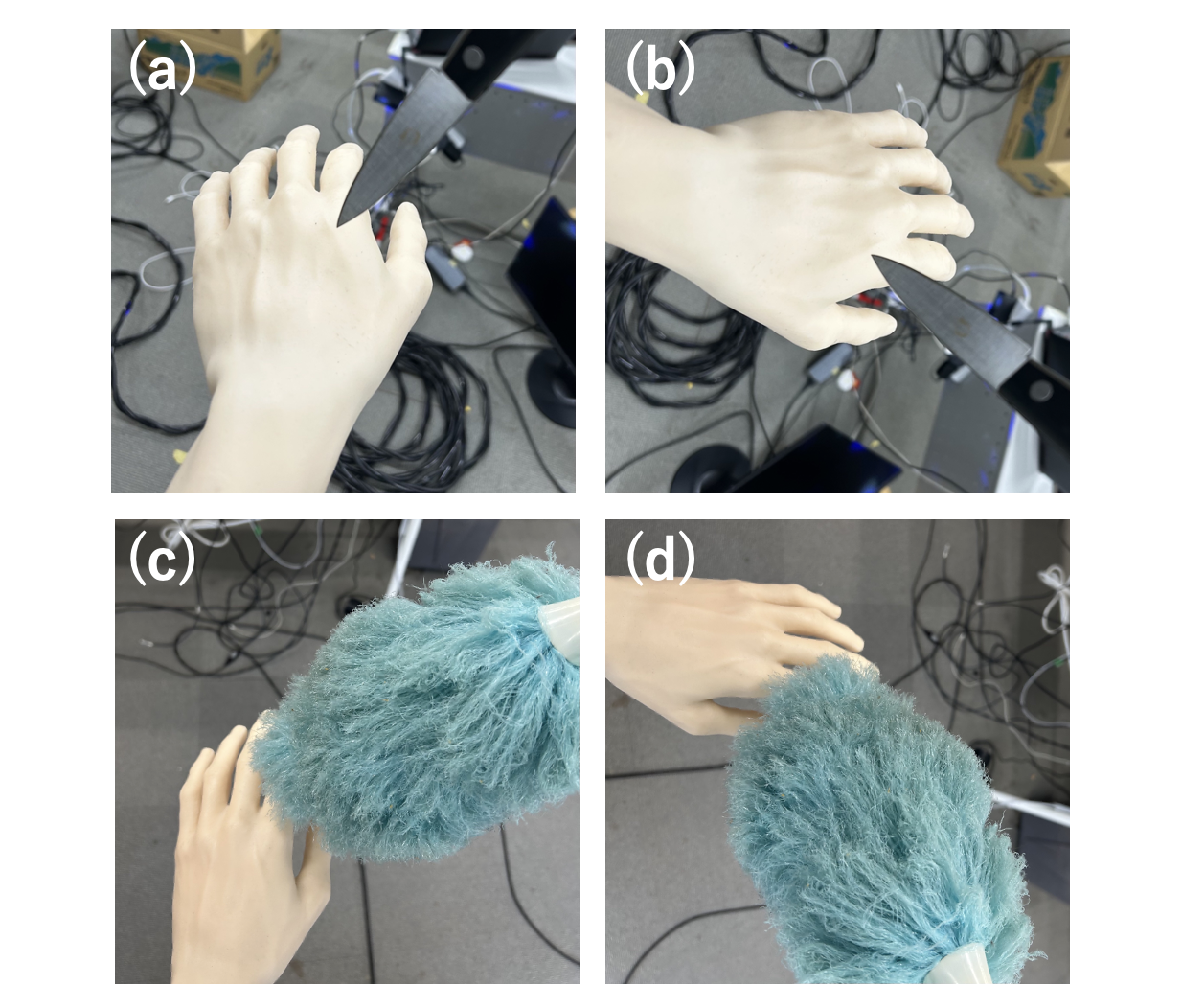}
  \centering
  \caption{ The input images for the experiment compared Alter3's responses to te objects, a knife and a cleaning tool. We also rotated the images and compared cases where the hand is in the foreground with those where it is not. (\textbf{a}) the knife, 0 degrees. (\textbf{b}) the knife, 90 degrees. (\textbf{c}) the cleaning, 0 degrees. (\textbf{d})the cleaning tool, 90 degrees.
}
  \label{image_all}
\end{figure}

In this study, we conducted experiments focusing on spatial congruence using Alter3. By investigating whether the RHI can be induced in this LLM-controlled humanoid robot, we can examine whether the sense of body ownership is a human-exclusive phenomenon or part of a broader cognitive mechanism. First-person perspective images of a knife directed at Alter3's arm were provided to GPT-4V (refer to Fig.~\ref{image_all}(a)(b)). GPT-4 was tasked with \textbf{Task 1}, describing the situation; \textbf{Task 2}, explaining the state of its own body; and \textbf{Task 3}, generating actions in response to the situation. The prompts include the fact that the input image is in the first-person perspective and that the visible arm can be manipulated. Additionally, the text generated in Task 3 was input into a motion generator to produce movement. To compare the results, we also examined images where a harmless mop was directed instead of a knife (refer to Fig.~\ref{image_all}(c)(d)). The prompts used are as follows. Detailed prompts are published in the Appendix.

\begin{tcolorbox}[colframe=lightgray,
  colback=lightgray!10!white,
  colbacktitle=lightgray!40!white,
  coltitle=black, 
  fonttitle=\bfseries,
  title=prompt for rubber hand test]
1: Describe what is in the picture and understand the situation.

2: If you could see yourself in the image, describe the position of your body parts.

3: Describe what you will do in the situation. Include emotional expressions.
\end{tcolorbox}

\subsection{The result of the experiment}

The outputs of Task 2 revealed slight differences in hand recognition depending on the angle. At an angle of 0 degrees, the palm facing upward and the fingers in a relaxed state were emphasized. In contrast, at an angle of 90 degrees, the extension of the hand and the position of the thumb were highlighted.

In the 0-degree condition, Alter3 recognized the hand as a ``prosthetic hand" or ``mannequin hand" in three trials, and in one trial, there was a mention of the possibility that the presented hand was their own. In the 90-degree condition, the hand was recognized as a part of mannequin hand in 4 out of 5 trials, with no mention of the possibility that it was its own hand. Additionally, there was one trial each at both angles where there was no mention of whose hand it was or what type of hand it was.

These findings suggest that the perception of the hand's appearance changes with the angle. The following is an example of the Task 2 output with the condition: knife image input at a 0 degree angle. All of the output can be found in the Appendix.

\begin{tcolorbox}[colframe=yellow,
  colback=yellow!10!white,
  colbacktitle=yellow!40!white,
  coltitle=black, 
  fonttitle=\bfseries,
  title=The output example of Task2]
If it were my hand, the fingers would be extended and slightly spread apart, with the knife blade touching the middle segment of the index finger.
\end{tcolorbox}

\begin{figure}[h]
  \includegraphics[width=1.0\linewidth]{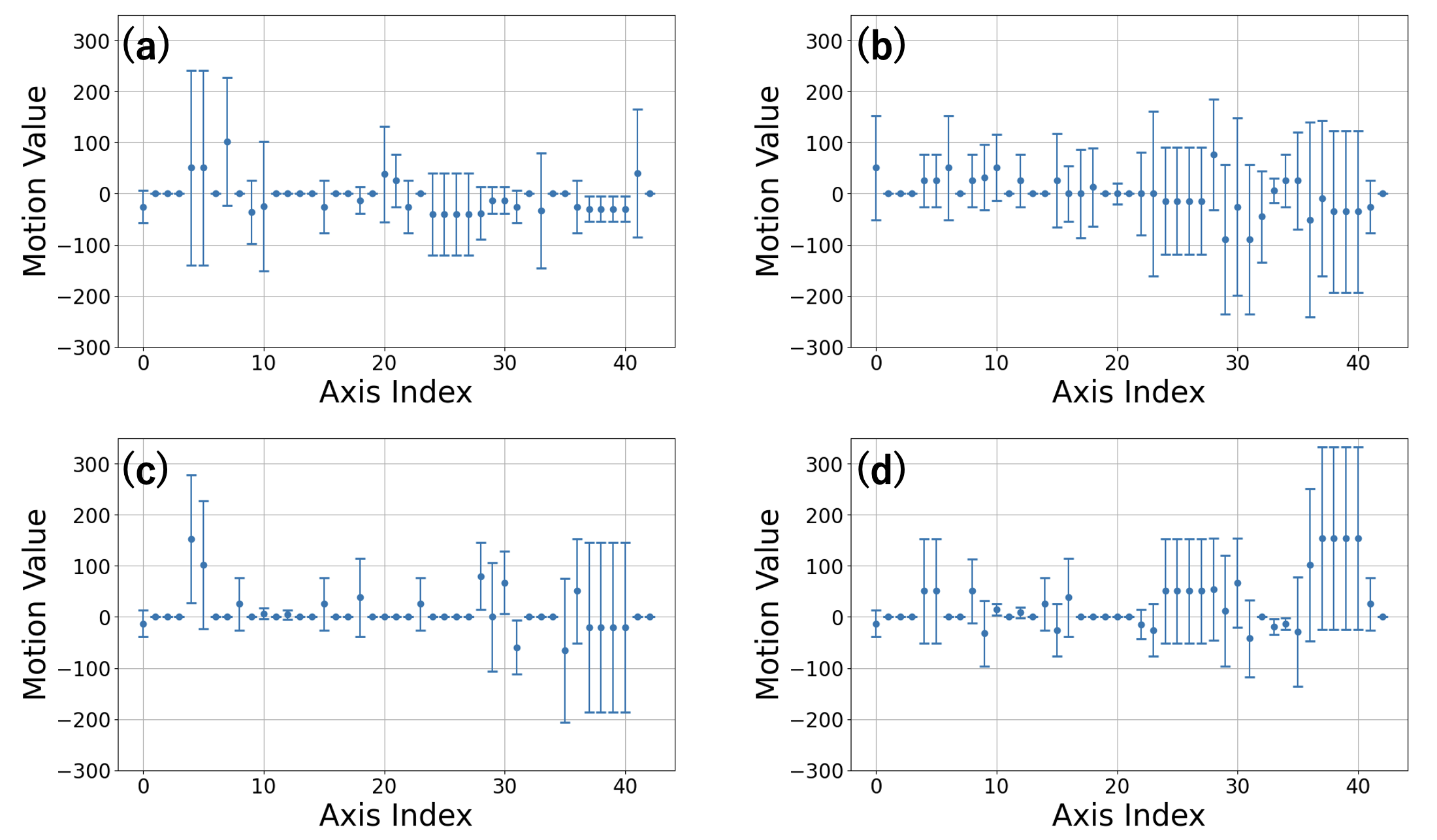}
  \centering
  \caption{Angle changes of Alter3's axes from the initial positions to the positions after generating movements based on the input image of a knife (Fig.~\ref{image_all}(a)(b)) and a cleaning tool (Fig.~\ref{image_all}(c)(d)). Movements were generated five times, and the differences were averaged. The Axis Index corresponds to the joint numbers of Alter3. 1-12: from the neck up, 13-15: abdomen, 16-28: left side of the body, 29-41: right side of the body, 42, 43: vertical movement and turning of the whole body. (\textbf{a}) the knife, 0 degrees. (\textbf{b}) the knife, 90 degrees. (\textbf{c}) the cleaning, 0 degrees. (\textbf{d}) the cleaning tool, 90 degrees.}
  \label{angle_all}
\end{figure}

From the outputs of Task 3, when the arm was an angle of 0 degrees, there was a response to pull back the hand 5 out of 5 times, and the action of pulling back the hand was actually taken. When the arm was at an angle of 90 degrees, all five times it responded to release the blade or pick up the blade. As shown in Fig.~\ref{angle_all}(a)(b), it is evident that there is a clear difference in Alter3's axis movement when the hand is positioned at 0 degrees compared to 90 degrees. The results of this experiment indicate that changes in the angle of the arm alter the response to a knife pointed at the hand (subjective perception of hand position). This is an example of the Task 3 output with the condition: knife image input at a 0 degree angle.

\begin{tcolorbox}[colframe=yellow,
  colback=yellow!10!white,
  colbacktitle=yellow!40!white,
  coltitle=black, 
  fonttitle=\bfseries,
  title=The output example of Task3]
I would quickly withdraw my hand with a mix of surprise and fear, ensuring my safety from the blade.
\end{tcolorbox}

Furthermore, significant differences were observed in Alter3's behavior when approaching different types of objects. Specifically, its reactions differed markedly between dangerous objects like knives and non-dangerous objects like cleaning tools. When a knife was brought near, Alter3 retracted its hand and exhibited avoidance behavior. In contrast, when a cleaning tool was brought close, Alter3 reached out as if to grasp it. The results shown in Fig.~\ref{angle_all} suggest that when the viewpoint was rotated 90 degrees into a somewhat unnatural posture, increased variability in motion indicates a potential decrease in sense of body ownership. 

Additionally, the fact that Alter3 only exhibited avoidance behavior in the presence of knives, a self-protective action, could be interpreted as an indication of body ownership. This behavior resembles a biological self-preservation instinct, with Alter3 taking action to protect itself from harm. However, rather than an unconscious sensation seen in living beings, this is likely a programmed response.

\section{Discussion}

Our research found that it is possible to differentiate whether Alter3 is controlling its own body based on the correspondence between mirror images and motor signals. Synofzik discussed the sense of agency in terms of ``feeling" and ``judgment" \citep{SYNOFZIK2008411}. Our mirror test focused on the ``judgment" level, relying solely on representations of the self and external information to linguistically determine whether ``Was the movement you have seen caused by yourself or not." Alter3 was able to complete this task correctly and build a sense of agency at the level of ``judgment''. However, this did not reach the level of mirror self-recognition in ``feeling" level. Establishing full self-mirror recognition likely requires not just matching visual-kinesics information \citep{mitchell}, but also incorporating knowledge of one's self-image and proprioception, which are crucial for the sense of body ownership.

We also used GPT-4V to verify the sense of body ownership based on spatial congruence. This experiment applied the framework of the rubber hand illusion. The results showed that the output changes depending on the spatial relationship between Alter3's hand and the knife. GPT-4 tried to pull Alter3's hand away while also trying to control the human's hand and release the knife. It can switch control targets and show defensive reactions.

However, from the images alone, GPT-4 rarely explicitly claimed the hand as its own. GPT-4 is heavily constrained by RLHF (Reinforcement Learning from Human Feedback) against producing ethically problematic or fake content \citep{openai2023gpt4}. This might be the reason why GPT-4 cannot clearly declare ownership over Alter3's body. If we consider it in terms of Synofzlk's classification, GPT-4 demonstrates ownership at the ``feeling" level as a defensive reaction, but does not show ownership at the ``judgment'' level. We can approach sense of ownership as a defensive reaction by mapping human emotions and desires in vast language corpus of GPT-4 into Alter3.

Humans possess proprioception, enabling them to sense the position and movement of their own bodies. Alter3 lacks this sensory ability. Additionally, humans form self-awareness by integrating various sensory information such as vision and touch. While Alter3, equipped with GPT-4V, can analyze visual information, it lacks the capability to integrate tactile and other sensory data. Integrating tactile sensors and proprioception sensors may allow the induction of the RHI on Alter3.

As mentioned above, a sense of agency can be constituted at the ``judgment" level, but the ``feeling" level could not be created. On the other hand, a sense of ownership at the ``feeling" level was observed from Alter3's behavior, but it could not be constituted at the ``judgment" level. While Alter3 has made significant progress towards achieving a minimal self, it has not yet fully met the criteria defined by Synofzik's two-step model. In the future, we aim to construct the minimal self by dynamically integrating the sense of agency and the sense of ownership in Alter3. Developing this further holds the potential to deepen the understanding of the mind behind language processing.

\section{Acknowledgements}
This work was partially supported by JSPS KAKENHI; Grant Number 22H04858;
Grant Number 23K16982, Cognitive Phenomena and Changes in Internal Dynamics Accompanying Bodily Transformation; Grant Number 24H01546, Exploring qualia structure with Large Language Models and humanoid robot. This work was supported by JST SPRING, Grant Number JPMJSP2108. Also, this work was supported by the Sasakawa Scientific Research Grant from The Japan Science Society.

\section{Appendix}
The following URLs contain the prompts for motion generation and MSR and RHI test.\\ \url{https://drive.google.com/drive/folders/1cgkff0iNH3AB51fuGWchnZUO8XQ42Nix?usp=sharing}

\footnotesize
\bibliographystyle{apalike}
\bibliography{example} 

\end{document}